\documentclass[conference]{IEEEtran}

\usepackage{hyperref}
\usepackage{url}
\usepackage{natbib}
\usepackage{array}
\usepackage{makecell} 
\usepackage{multirow}
\usepackage{graphicx}
\usepackage{tikz}  
\usepackage{cleveref}
\usepackage{acronym}
\usepackage{subcaption}
\acrodef{nlp}[NLP]{Natural Language Processing}
\acrodef{lm}[LM]{Language Model}
\acrodef{wsd}[WSD]{Word Sense Disambiguation}
\acrodef{lane}[LANE]{Lexical Adversarial Negative Example}
\acrodef{lscd}[LSCD]{Lexical Semantic Change Detection}
\acrodef{oewn}[OEWN]{Open English WordNet}

\title{Adverbs Revisited: Enhancing WordNet Coverage of Adverbs with a Supersense Taxonomy}

\author{\IEEEauthorblockN{Jooyoung Lee}
\IEEEauthorblockA{\textit{Department of Linguistics}\\
\textit{Brown University}\\
  Providence, RI 02912, United States}
\and
\IEEEauthorblockN{Jader Martins Camboim de S\'a}
\IEEEauthorblockA{\textit{Faculty of Science Technology and Medicine} \\
\textit{University of Luxembourg}\\
5 avenue des Hauts-Fourneaux, L-4362, Esch-sur-Alzette, Luxembourg\\
first.second@list.lu}
}

\begin{document}
\maketitle

\begin{abstract}
WordNet offers rich supersense hierarchies for nouns and verbs, yet adverbs remain underdeveloped, lacking a systematic semantic classification. We introduce a linguistically grounded supersense typology for adverbs, empirically validated through annotation, that captures major semantic domains including manner, temporal, frequency, degree, domain, speaker-oriented, and subject-oriented functions. Results from a pilot annotation study demonstrate that these categories provide broad coverage of adverbs in natural text and can be reliably assigned by human annotators. Incorporating this typology extends WordNet’s coverage, aligns it more closely with linguistic theory, and facilitates downstream NLP applications such as word sense disambiguation, event extraction, sentiment analysis, and discourse modeling. We present the proposed supersense categories, annotation outcomes, and directions for future work.
\end{abstract}

\section{Introduction}
\label{sec:intro}

As a primary lexical class, adverbs perform a range of semantic functions, from answering fundamental questions about an event, such as how it was performed (manner), when it occurred (temporal), or to what extent a property holds (degree), to expressing speaker attitude, discourse stance, and logical relations between propositions. Despite this semantic richness, adverbs have long occupied an ambiguous and often marginalized position in linguistic classification, frequently described as a ''residual'' or ''wastebasket'' category \cite{Huddleston2002TheCG, nikolaev2023adverbs}. Words are often assigned to this category not because they share definable grammatical properties, but because they fail to conform to the morphological and syntactic criteria of nouns, verbs, adjectives, prepositions, or conjunctions. This stands in contrast to the detailed formal and semantic analyses developed for nouns and verbs. \cite{conlon1994adverbial}.

\begin{figure}[htpb]
\centering
\begin{tikzpicture}[every node/.style={align=center}]
\node (root) at (0, -1) {\textbf{adverb}};
\node (subj) at (-2, 0) {\textbf{subject-oriented}\\ \textit{Stupidly}, he lied.};
\node (speaker) at (-2.75, -2.0) {\textbf{speaker-oriented}\\\textit{Unfortunately}, no.};
\node (conjunctive) at (-2.0, -2.95) {\textbf{conjunctive}\\Yes, \textit{so} are you.};
\node (degree) at (-2.75, -1.0) {\textbf{degree}\\It's \textit{insanely} cold here.};
\node (manner) at (1.5, 0) {\textbf{manner}\\ He walked \textit{fast}.};
\node (domain) at (-0.5, 1.0) {\textbf{domain}\\ \textit{Technically}, she is right.};
\node (freq) at (0.5, -3.5) {\textbf{frequency}\\ I go to school \textit{daily}.};
\node (spatial) at (2.2, -2.5) {\textbf{spatial}\\ It is \textit{there}.};
\node (temporal) at (2.2, -1.0) {\textbf{temporal}\\ \textit{Before} that.};

\draw[->] (subj) -- (root);
\draw[->] (speaker) -- (root);
\draw[->] (conjunctive) -- (root);
\draw[->] (manner) -- (root);
\draw[->] (freq) -- (root);
\draw[->] (spatial) -- (root);
\draw[->] (degree) -- (root);
\draw[->] (temporal) -- (root);
\draw[->] (domain) -- (root);
\end{tikzpicture}
\caption{Taxonomy for adverb supersenses.}
\label{fig:taxonomy}
\end{figure}
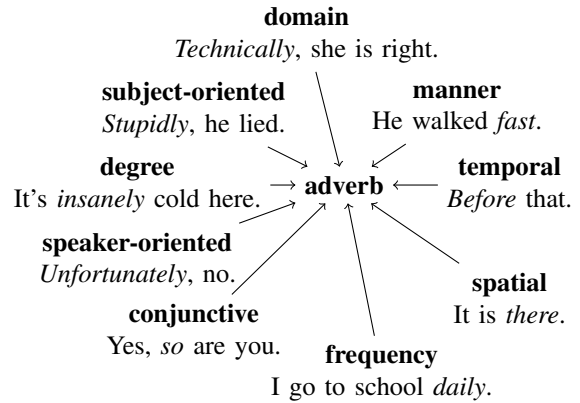

The heterogeneity of adverbs arises from the multiple dimensions along which they vary. Syntactically, adverbs can attach at different heights, within the VP, clause, or discourse level, modifying events, participants, or propositions. Semantically, they range from event-oriented (\textit{quickly, carefully}) to epistemic and evaluative (\textit{probably, fortunately}) to domain-restrictive (\textit{legally, technically}) meanings. Morphologically, English adverbs derive through diverse pathways, productively with \textit{-ly}, through lexicalization (\textit{often, maybe}), or borrowing (\textit{a priori, per se}), each contributing to internal diversity. Collapsing such functionally distinct elements under a single label produces an uninformative and theoretically incoherent grouping that obscures meaningful semantic distinctions.

This conceptual heterogeneity has far-reaching implications beyond linguistic theory. In computational linguistics and large lexical databases, adverbs often receive limited or inconsistent representation, reflecting their ambiguous grammatical status. Major resources such as WordNet, for instance, tend to prioritize manner adverbs while overlooking domain, epistemic, or evaluative meanings. This imbalance illustrates how theoretical uncertainty surrounding adverbs has translated into gaps in computational modeling and lexical coverage, motivating the need for a more systematic and semantically grounded classification.

Distinguishing these meanings is not merely a matter of theoretical precision. Adverbs frequently serve as indicators of how speakers assess states of affairs, making them highly informative for computational tasks such as opinion mining, stance detection, and sentiment analysis \cite{Pang2008OpinionMA}. Their semantic function extends beyond simple modification to include attitudinal and intensificational meanings that shape pragmatic interpretation. Without an explicit taxonomy capturing distinctions such as manner, evaluative, and intensifier uses, NLP systems cannot systematically model these meanings. 

This paper addresses this gap by proposing a linguistically motivated and empirically validated typology of adverbs and demonstrating its effectiveness through computational classification. We present a taxonomy encompassing event-related, domain-related, subject-oriented, and speaker-oriented adverbs (see Cref{fig:taxonomy}), and implement an annotation pipeline to facilitate the systematic expansion and annotation of WordNet adverbs.

\section{Background}
\label{sec:background}
In most syntactic and computational frameworks, adverbs have been treated as adjuncts, optional modifiers situated outside the core predicate–argument structure. Because early parsing systems were primarily designed to capture argument relations, adverbial phrases were typically attached in a flat, non-hierarchical fashion, with little representation of their semantic contribution. This simplified treatment, inherited by subsequent treebanks and dependency parsers, has had enduring effects: adverbs are encoded as syntactically peripheral and semantically underspecified elements \cite{ernst2002syntax, mcconnell1982adverbs}.

These structural simplifications have shaped how adverbs are represented in modern computational resources and tasks. Large lexical databases such as WordNet \cite{Miller1995WordNetAL, Fellbaum1998CombiningLC}, syntactic annotation schemes like the Penn Treebank \cite{mitchell1993penntreebank}, and semantic frameworks including semantic role labeling all encode adverbs in simplified or undifferentiated ways, reflecting a broader trend of structural and semantic underrepresentation across NLP.

\subsection{WordNet}
WordNet is among the most influential lexical–semantic resources in computational linguistics, \ac{nlp}, and cognitive science. Its architecture organizes words into synsets grouped by supersenses, capturing broad conceptual domains across parts of speech. It covers 26 noun supersenses and 15 verb supersenses, allowing for systematic semantic generalization. On the other hand, all adverbs are placed under a single supersense, ''adv.all'', which collapses their internal diversity into a monolithic category. This flattening of the semantic space not only underrepresents the heterogeneity of adverbial meaning but also constrains the usefulness of WordNet for adverb-related research.

The WordNet documentation explicitly states that ''there are only few adverbs in WordNet … as the majority of English adverbs are straightforwardly derived from adjectives via morphological affixation'' \cite{WordNet_website}. This assumption simplifies adverb formation to a purely morphological process and conflates morphological derivation with semantic equivalence, failing to recognize the interpretive autonomy of adverbs. For instance, while the adverb \textit{stupidly} in \textit{He danced stupidly} concerns the manner in which John danced, the counterpart in \textit{Stupidly, he danced} concerns the speaker's evaluation on John's decision to dance and does not have any inference about \textit{how} he danced. Such semantic contrast cannot be predicted from adjectival meaning alone. The assumption that adverb formation is semantically straightforward is therefore untenable. 

\subsection{Penn Treebank}
The Penn Treebank (PTB), a foundational syntactic resource for \ac{nlp}, serves as the basis for countless parsing models and downstream datasets. However, its treatment of adverbs mirrors the structural simplifications seen in WordNet. The tagset distinguishes only four adverb categories: RB for general adverbs such as \textit{quickly}, \textit{often}, and \textit{perhaps}; RBR for comparative adverbs such as faster and earlier; RBS for superlative adverbs such as best and fastest; and WRB for wh-adverbs such as \textit{when, where,} and \textit{why}. These tags primarily capture morphological and comparative information, making no reference to syntactic scope or semantic function. As a consequence, adverbs with vastly different meanings and syntactic behaviors such as \textit{quickly} (manner), \textit{probably} (epistemic), \textit{fortunately} (speaker-oriented), and \textit{technically} (domain) are all collapsed into the identical RB tag, treating them as a single class.

This coarse-grained tagging is not merely a semantic omission; it creates a problem for syntactic parsing itself. The primary challenge a parser faces with an RB tag is attachment ambiguity. It has no information to determine if the adverb should attach low to the VP (e.g., \textit{He spoke clearly} in a manner sense) or high to the entire Sentence (e.g., \textit{Clearly, he is joking} in a speaker-oriented sense). By lumping all these adverbs together, the tagset discards the single most important clue for resolving this ambiguity. Since an adverb’s interpretive role is inseparable from its syntactic scope, a semantically grounded taxonomy is not merely desirable but necessary for improving both syntactic and semantic accuracy.

\subsection{PropBank}
The Proposition Bank (PropBank; \citealt{palmer2005proposition}) extends the Penn Treebank by adding a layer of semantic role labeling, enriching syntactic parses with predicate–argument structure. Its design represents a significant step forward: PropBank introduced a detailed inventory of adjunct types that partially disaggregate adverbial meaning, including labels such as ARGM-TMP (temporal), ARGM-LOC (locative), ARGM-CAU (causal), ARGM-PRP (purpose), ARGM-MNR (manner), and ARGM-DIS (discourse connective). These distinctions capture coarse semantic relations that the PTB’s RB tag entirely ignores, allowing for more interpretable role structures and finer-grained feature extraction for semantic parsing models.

However, despite these improvements, PropBank’s treatment of adverbs still relies on a catch-all category, ARGM-ADV, which functions as a repository for any adverb that does not fit the existing labels. As Nikolaev (2023) \cite{nikolaev2023adverbs} observe, this design reproduces the ''wastebasket'': multiple distinct semantic types including epistemic, evaluative, subject-oriented, and domain adverbs—are merged into one ambiguous slot. A classic example like \textit{stupidly} in \textit{Stupidly, he answered} (meaning it was stupid of him) is neither a simple manner nor a speaker evaluation. With no dedicated category, it is inevitably forced into one of these ill-fitting boxes, most likely ARGM-MNR, thus demonstrating that PropBank's schema is blind to this core linguistic distinction.

\subsection{NLP tasks}

The simplified representation of adverbs in foundational resources such as WordNet, the Penn Treebank, and PropBank has cascading effects across NLP tasks, causing a systematic loss of semantic precision. By treating adverbs as a single or weakly differentiated category, these resources obscure distinctions in semantic scope, orientation, and function, differences that are crucial for interpreting event structure and speaker stance. This problem is especially clear in Word Sense Disambiguation, where the sense inventories themselves encode the bias. For instance, SenseBERT \cite{levine-etal-2020-sensebert} doesn't learn polysemy for adverbs and similarly WiC models \cite{pilehvar-camacho-collados-2019-wic} are not trained for adverb differentiation. Adverb are often merged into one entry, leaving systems unable to identify the correct sense because the relevant contrast is missing. Such limitations affect machine translation and sentiment analysis as well, where adverbs of stance, degree, and certainty are often mistranslated or misinterpreted. A semantically grounded taxonomy of adverbs would provide explicit cues for scope and orientation, improving semantic role identification.

This problem extends directly to Semantic Role Labeling (SRL), where the semantic layer in PropBank inherits the Treebank’s structural simplifications. Although PropBank improves on purely syntactic annotation by introducing a range of adjunct roles, its framework still conflates functionally distinct adverbs under general labels such as ARGM-ADV or ARGM-MNR. This design erases the distinction between adverbs that modify the event, those that evaluate the agent’s participation, and those that comment on the proposition or discourse. As a result, SRL models cannot exploit adverb semantics to constrain predicate–argument interpretation or infer pragmatic scope. A semantically grounded taxonomy of adverbs, which encodes the level of modification and orientation, would therefore provide crucial cues for both syntactic attachment and argument interpretation, enhancing the performance of SRL, WSD, and related tasks that depend on fine-grained semantic distinctions.

\section{Related Work}
\label{sec:related}
Developing semi-automatic methods for WordNet expansion remains a long-standing challenge \cite{ciaramita-johnson-2003-supersense, curran-2005-supersense}, since automatic approaches have yet to meet the quality standards expected of lexicographic resources \cite{rundell2024automating}.

Tsetkov \textit{et al.} (2014) \cite{tsvetkov-etal-2014-augmenting-english} address the absence of a semantic hierarchy for adjectives in English WordNet, a limitation closely parallel to that of adverbs. While WordNet provides a rich taxonomic structure for nouns and verbs, adjectives are organized only into flat, unstructured clusters. To overcome this limitation, the authors introduce a coarse-grained taxonomy of 13 adjective supersenses, adapted from GermaNet, thereby imposing a hierarchical organization onto the adjective lexicon. Acknowledging that fully manual annotation of all 18,156 adjective synsets would be infeasible, they develop a semi-automatic classification approach.

\cite{maziarz-etal-2016-adverbs} tackle the under-representation of adverbs in plWordNet, arguing that simply treating them as derivatives of adjectives is insufficient. Their methodology is twofold: first, they define a formal set of semantic relations for adverbs, including hyponymy, gradation, and antonymy, which is largely adapted from their adjective model. Second, they implement a semi-automatic procedure to bootstrap the adverb lexicon from the existing adjective network. This procedure generates adverbial counterparts for adjective lexical units and groups them into synsets that mirror the structure of the adjective synsets, providing a starting point for manual verification.

While previous work has investigated the expansion of supersenses for adjectives and the extension of adverb coverage in the Polish WordNet, to the best of our knowledge, no prior study has developed a taxonomy or systematically expanded the coverage of adverbs in the English WordNet.

\section{Taxonomy}
\label{sec:taxonomy}
The typology proposed here builds on the tripartite classification of manner, subject-oriented, and speaker-oriented adverbs \cite{jackendoff1972semantic}, which first recognized that adverbs differ in the level of meaning they target: the event, the agent, or the speaker, respectively. 

Building on Jackendoff’s tradition, our typology expands this inventory to include classes that systematically differ in interpretive scope, such as frequency, temporal, spatial, degree, domain, focus, and conjunctive adverbs. We ground this expanded classification in the rigorous syntax-semantics mapping proposed by Cinque (1999) \cite{cinque1999adverbs} and the scope-based approach of Ernst (2002) \cite{ernst2002syntax}. Both works argue that these distinct semantic functions are not random but correspond to a rigid hierarchy of syntactic positions, finding that is critical for resolving the attachment ambiguity that plagues computational parsers.

This combined theoretical grounding provides a principled framework for our classification. It allows us to move beyond a flat list and organize adverbs by their level of semantic scope: from event-internal modifiers (manner, spatial, temporal, frequency) and agent-evaluative modifiers (subject-oriented), to proposition-level modifiers (speaker-oriented, domain) and discourse-level operators (conjunctive, focus). 

\begin{itemize}
    \item \textbf{Manner} adverbs describe how an event or action is performed. They can typically be paraphrased using the expression \textit{in a [X] manner}. For example, \textit{He danced stupidly} may be rephrased as \textit{He danced in a stupid manner}.
    \item \textbf{Subject-oriented} adverbs attribute a property or attitude to the subject of the sentence in relation to the event. These adverbs often allow a paraphrase of the form \textit{It was [X] of [SUBJECT] to [VERB]}, as in \textit{Stupidly, he walked into traffic} → It was stupid of him to walk into traffic.
    \item \textbf{Speaker-oriented} adverbs express the speaker’s stance, evaluation, or attitude toward the proposition or the act of speaking. They can frequently be paraphrased with constructions such as \textit{I [believe/say/judge] that…} or \textit{It is [unfortunate/fortunate/evident] that…}. For instance, \textit{Presumably, he missed the deadline} corresponds to I presume that he missed the deadline, while Unfortunately, the project failed can be restated as It is unfortunate that the project failed, and Frankly, I disagree as I say this frankly.
    \item \textbf{Frequency} adverbs indicate how often an event occurs and typically answer the question \textit{How often?}, as in She frequently visits her grandmother. 
    \item \textbf{Temporal} adverbs specify when an event happens or its duration, answering questions like \textit{When?} or \textit{For how long?} (e.g., She arrived yesterday; He stayed briefly).
    \item \textbf{Spatial} (or locative) adverbs refer to where the event takes place and can be identified by the question \textit{Where?}, as in He stood outside.
    \item \textbf{Degree} adverbs describe the intensity or extent of another element, typically answering \textit{To what extent?} or \textit{How much?}, as in She is very happy.
    \item \textbf{Domain} adverbs restrict the interpretation of the proposition to a particular semantic or disciplinary domain, often paraphrasable as \textit{In a [X] sense} or \textit{From a [X] perspective}; for example, \textit{Politically, the policy is controversial} → \textit{In a political sense, the policy is controversial}.
    \item \textbf{Focus} adverbs highlight or limit the scope of the proposition to a specific constituent, marking inclusion, exclusion, or emphasis. They answer questions such as \textit{What part of the clause is being emphasized or limited?} or \textit{Does it indicate exclusivity, inclusion, or emphasis?}. Examples include: \textit{Only John passed the exam} (exclusivity); \textit{Even Mary noticed the mistake} (unexpected inclusion); \textit{She also attended the meeting} (addition); and \textit{He mainly spoke about politics} (restriction).
    \item \textbf{Conjunctive} adverbs signal opposition, correction, connection, or contrast between propositions or discourse segments, relating the current clause to an alternative or prior context. They are often paraphrasable by \textit{in contrast}, \textit{on the other hand}, or \textit{however}. Typical examples include \textit{However, she decided to stay} (contrast with a preceding statement), \textit{Conversely, rural areas saw a decline} (opposition), \textit{Instead, he chose to wait} (correction), and \textit{Nevertheless, they continued the project} (concession).
\end{itemize}

\section{Methodology}
\label{sec:methodology}
\begin{figure*}[htpb]
    \centering
    \includegraphics[width=\linewidth]{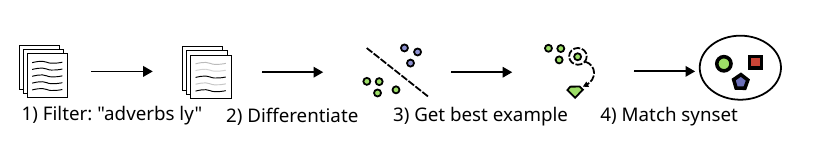}
    \caption{Pipeline for assisting the annotation of adverbs.}
    \label{fig:pipeline}
\end{figure*}

The expansion of adverb supersenses in \ac{oewn} follows a semi-automatic annotation pipeline, in which each computational stage is manually validated by professional linguists. The process first involves categorizing existing adverbs in \ac{oewn} to determine which semantic senses warrant inclusion in the extended taxonomy. New adverbial senses are then identified through large-scale corpus analysis, where adverbs are extracted, their contextual meanings distinguished, and representative examples selected. These examples are subsequently integrated into \ac{oewn} in accordance with its structural and formatting conventions. An overview of this workflow is presented in \Cref{fig:pipeline}.

\subsection{Adverb Extraction}
The first step in \Cref{fig:pipeline}, consists in collecting a large, representative corpus and extracting all adverbial forms. Adverb candidates are identified through a combination of dictionary lookup and morphological detection, primarily targeting common adverbial suffixes (e.g., \textit{-ly}). Surface forms are preserved throughout the process to maintain alignment with corpus usage. Ambiguous forms that can function as multiple parts of speech (e.g., \textit{fast, hard}) are excluded at this stage to avoid introducing noise into the adverb inventory.

\subsection{Lexical Sense Differentiation}
For each extracted adverb, \Cref{fig:pipeline} second step, distinct lexical usages are identified using cosine distance of the contextualized embeddings of each usage. The model compares contextual embeddings across occurrences of the same adverb, computing pairwise cosine distances to cluster similar contexts. New sense discovery is entirely data-driven, with the number of senses emerging from the clustering process rather than being predefined. Following the \ac{oewn} guidelines, a cluster should be composed of at least 100 elements to be considered a valid sense \cite{mccrae-etal-2021-globalwordnet}. Each resulting sense is subsequently assigned a dictionary definition by annotators, who also classify the adverb according to its semantic category (e.g., manner, degree, temporal, frequency).

\subsection{Example Sentence Selection}
To provide representative examples for each sense, sentences are ranked according to the contextual embedding uncertainty estimated by the differentiation model. The sentence with the lowest entropy value is selected as the prototypical example. Sentences are further filtered by length, retaining only those containing between 3 and 20 tokens to ensure naturalness and readability. Annotators then manually verify each selected example to confirm that it accurately illustrates the intended sense.

\subsection{Synset Formation}
Synset construction is performed by aligning newly identified adverb senses with existing or newly created adverb synsets. Sense alignment is based on cosine similarity between sense embeddings, applying a similarity threshold to determine equivalence. Only adverbs belonging to the same semantic category are eligible for grouping. Existing \ac{oewn} resources are consulted to validate and anchor new synsets within the current lexical network. When multiple senses show near-equivalence or redundancy, they are manually reviewed and merged as appropriate.

\subsection{Validation and Integration}
All newly generated entries are formatted in YAML, following the \ac{oewn} schema. The final output includes the adverb lemma, sense identifier, gloss, and verified example sentence. Inter-annotator agreement is computed for the taxonomy for the initial round, then each annotator works on independent examples. Finally, at the synset construction, a manual post-processing step ensures consistency by removing or merging near-duplicate synsets before integration into \ac{oewn}.

\section{Experiments}
\label{sec:experiments}
To improve the adverbs coverage in \ac{oewn} \cite{globalwordnet} the open version derived from WordNet \cite{wordnet}, we start by training two annotators in the taxonomy mentioned in \Cref{sec:taxonomy} and detailed in the Appendix. These annotators have English as their second language and a background in linguistics. After some short rounds to train and clear annotators doubts, we run a full annotation round. We observed a substantial annotator agreement (Cohen's kappa = 0.67; n = 229).

\begin{figure}[htpb]
    \centering
    \includegraphics[width=\linewidth]{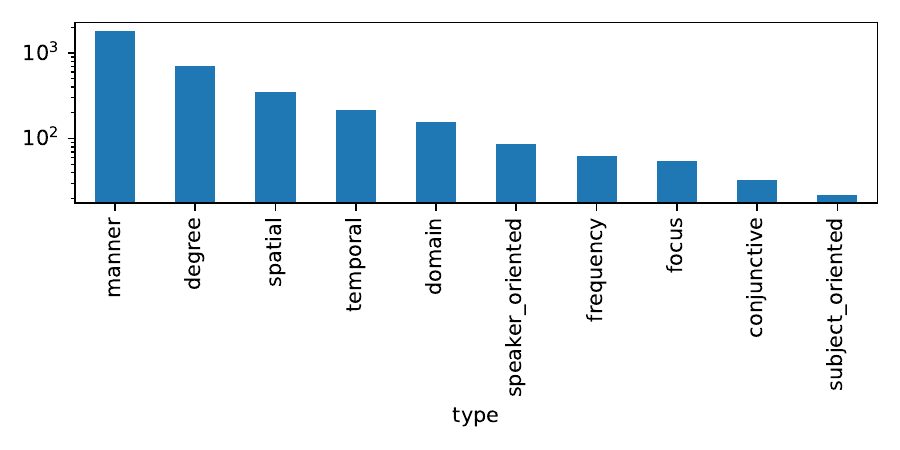}
    \caption{Counts of types for adverbs in WordNet.}
    \label{fig:count_wordnet}
\end{figure}

After the initial round, the annotators independently annotate examples from WordNet following the guidelines, in \Cref{tab:adverb_examples} we demonstrate examples of the supersenses annotated for existing examples in \ac{oewn}. In \Cref{fig:count_wordnet} we plot the distribution of categories for adverbs after this annotation.

\begin{table*}[htpb]
\centering
\begin{tabular}{llll}
\hline
\textbf{Category} & \textbf{Synset ID} & \textbf{Adverb} & \textbf{Usage} \\
\hline
conjunctive & 00043413-r & thus & it is late and thus we must go \\
degree & 00006640-r & significantly & our budget will be significantly affected [...] \\
domain & 00131423-r & semantically & semantically empty messages \\
focus & 00009062-r & alone & [...] rests on the prosecution alone \\
frequency & 00256795-r & biannually & we hold our big sale biannually \\
manner & 00248938-r & horrifyingly & he laughed horrifyingly \\
spatial & 00259792-r & inland & the town is five miles inland \\
speaker\_oriented & 00201575-r & hopefully & hopefully the weather will be fine on Sunday \\
subject\_oriented & 00473325-r & superstitiously & superstitiously he refused to travel on Friday [...] \\ 	
temporal & 00061170-r & previously & he was previously president of a bank \\
\hline
\end{tabular}
\caption{Examples of adverb categories, their WordNet synset IDs, and usage examples.}
\label{tab:adverb_examples}
\end{table*}

To extend existing usages for adverbs and discover new ones., we began by compiling a large-scale corpora. For each adverb, we extracted millions of usage instances from OpenSubtitles \cite{opensubtitles} and Fineweb-edu \cite{fineweb_edu}.
To ensure linguistic diversity, we included data from multiple domains and registers. Each instance was stored along with its surrounding sentence or paragraph context to support later sense clustering and disambiguation.

To capture the potential polysemy of each adverb, we organized its usage instances into clusters representing distinct contextual senses. This step was performed using XL-Lexeme \cite{cassotti-etal-2023-xl}, available in HuggingFace\footnote{\url{https://huggingface.co/pierluigic/xl-lexeme}}. XL-Lexeme, a multi-lingual sense differentiation with empirical good generalization, computes contextual embeddings for each adverb occurrence and then we group them using a Agglomerative Clustering with a threshold of 0.4. Each resulting cluster corresponds to a potential sense of the adverb, defined by semantically coherent usage contexts.

After the supersense classification, we used the same LLM to generate sense definitions for each cluster. Each definition was produced by prompting the model with representative examples from the cluster and requesting a concise, WordNet-style definition. We selected the examples with the lowest entropy when the adverb is masked and feed to the XLM-RoBERTa model. This process allowed us to automatically select high-quality examples and derive lexicographically interpretable sense entries for each adverb.

All automatically generated senses underwent manual evaluation by trained annotators. The evaluation involved verifying the accuracy of the supersense label, checking for redundancy across clusters, and refining the sense glosses for lexical and stylistic consistency. To this task we assigned two annotators where English is their second language. In the next section we report inter-annotator agreement.
Errors were corrected, redundant clusters were merged, and unclear definitions were rewritten following WordNet conventions. The final resource includes 418 new adverbs or usages not present in WordNet and constitutes a validated extension of adverb supersenses aligned with the WordNet ontology.

\begin{figure}
    \centering
    \includegraphics[width=\linewidth]{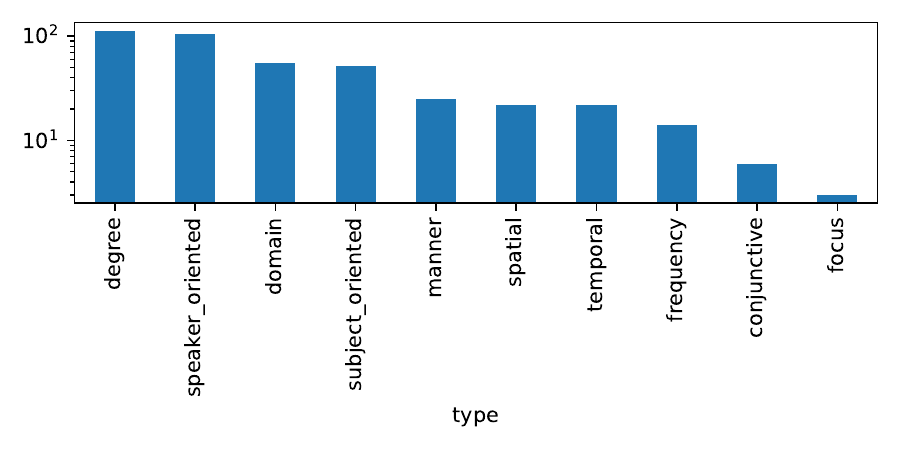}
    \caption{Counts of types for adverbs not present in WordNet.}
    \label{fig:count_new}
\end{figure}

We discarded existing senses based on the adverb category and in \Cref{fig:count_new} we present the statistics of the new obtained category and senses.

The resulting resource, including the adverb sense inventory and annotation guidelines, will be released under an open license and made available via Github.

\section{Discussion}
\label{sec:discussion}
The primary objective of this work was to expand the coverage of adverbs in the \ac{oewn} through a novel semi-automatic pipeline to assist professional annotators. The methodology, combining large-scale computational analysis with essential manual validation, proved effective, resulting in the creation of 418 new adverb synsets previously absent from WordNet. This outcome validates the hybrid \textit{human-in-the-loop} approach, demonstrating that modern NLP tools can significantly accelerate and scale lexicographic efforts.

A key strength of our pipeline is the integration of large models (LLMs) for complex annotation tasks. This allowed our trained annotators to focus on validation and refinement rather than from scratch sense discovery. The selection of large, diverse corpora ensured that the identified senses reflect contemporary and varied language use.

The validation phase yielded a substantial inter-annotator agreement (Cohen's Kappa = 0.67) for the semantic category classification. This result underscores the inherent ambiguity and complexity of adverb classification, even for trained annotators working with detailed guidelines. It also confirms that the task is feasible and that our guidelines provide a solid basis for consistent annotation. The manual verification by two trained L2 English annotators was crucial for correcting errors, merging redundant clusters, and ensuring the final definitions met \ac{oewn}'s lexicographical standards. The comparison between the original adverb distribution in \Cref{fig:count_wordnet} and our new additions in \Cref{fig:count_new} highlights that our method successfully identified senses in categories that were previously underrepresented.

\section{Conclusion}
\label{sec:conclusion}

This paper presented a semi-automatic pipeline for the identification, differentiation, and integration of new adverb synsets into \ac{oewn}. By combining contextualized embeddings for sense clustering, LLMs for bootstrapped annotation, and expert manual validation, we added 418 new high-quality adverbial senses and categorized 3,493 existing senses in \ac{oewn}. This work makes two key contributions: first, it delivers a substantial and concrete expansion of \ac{oewn}, an essential lexical resource for the NLP community; second, it introduces a replicable methodology that effectively balances the scalability of computational models with the precision of human annotation.

The resulting adverb inventory and accompanying annotation guidelines, which will be released publicly, constitute a valuable resource for future research in lexical semantics, computational lexicography, and natural language understanding. Future work should extend this pipeline to address POS-ambiguous adverbs excluded from the present study and to evaluate the transferability of this methodology to other parts of speech whose sense inventories remain underspecified or outdated.

\section*{Limitations}
\label{sec:limitations}
Despite the pipeline’s success, several limitations warrant acknowledgment. First, the methodology excluded ambiguous forms that can function across multiple parts of speech (e.g., fast, hard). While this restriction was necessary to minimize noise during extraction and clustering, it leaves both our resource and \ac{oewn} with limited coverage of morphosyntactically ambiguous adverbs, which are among the most frequent and semantically complex in English. Addressing these cases remains an important direction for future work. Second, the quality of the automated components depends on model-specific factors—particularly the XL-Lexeme embedding model and the clustering threshold (set to 0.4). Alternative models or parameters could yield different levels of sense granularity, and the need for manual post-merging indicates that clustering occasionally produced redundant or overly fine-grained senses. Finally, although two annotators carried out the validation phase, incorporating a larger and more diverse annotation pool, including L1 English-speaking lexicographers, would likely improve consistency and further enhance the robustness of the resulting resource.


\section{Bibliographical References}
\bibliographystyle{plain}
\bibliography{anthology,custom,languageresource}

\end{document}